\documentclass[letterpaper, 10 pt, conference]{ieeeconf}  % Comment this line out if you need a4paper

\IEEEoverridecommandlockouts                              % This command is only needed if 
\usepackage{times}

\usepackage{caption}
\usepackage{multicol}
\usepackage[bookmarks=true]{hyperref}
\usepackage{amsmath, amssymb}
\usepackage{booktabs}
\usepackage{graphicx}
\usepackage{xcolor}
\usepackage{float}
\usepackage{mathrsfs} %
\usepackage{caption}
\usepackage{tabularray} 
\usepackage{tabularx} % for 'tabularx' env. and 'X' col. type
\usepackage{ragged2e} % for \RaggedRight macro
\usepackage{cuted}
\usepackage{array}      % 
\usepackage{multirow}   % 
\usepackage[utf8]{inputenc}
\usepackage{svg} 
\usepackage{algpseudocode}  
\usepackage{pifont}
\usepackage{amsfonts}
\usepackage{dsfont}
\usepackage{balance}
\usepackage{stfloats}
\hypersetup{hidelinks} 
\usepackage{balance}
\usepackage[
    style=ieee,
    natbib=true,
    citestyle=numeric-comp,
    doi=false,
    isbn=false,
    url=false]{biblatex} 

\makeatletter
\@ifundefined{labelindent}{}{%
  \let\labelindent\relax
}
\makeatother
\usepackage{enumitem}

\makeatletter
\let\NAT@parse\undefined
\makeatother

\usepackage{cleveref}

\definecolor{BrickRed}{rgb}{.72,0,0}

\title{\LARGE \bf
Ego-Vision World Model for Humanoid Contact Planning
}

\IEEEaftertitletext{\vspace{-2.0\baselineskip}} 

\author{
    Hang Liu\textsuperscript{2}, 
    Yuman Gao$^{1}$, 
    Sangli Teng\textsuperscript{1},
    Yufeng Chi\textsuperscript{1}, \\
    Yakun Sophia Shao\textsuperscript{1}, 
    Zhongyu Li\textsuperscript{3}, 
    Maani Ghaffari\textsuperscript{2}, and
    Koushil Sreenath\textsuperscript{1} \\
    \thanks{$^{1}$University of California, Berkeley}
    \thanks{$^{2}$University of Michigan, Ann Arbor}
    \thanks{$^{3}$The Chinese University of Hong Kong}
}

\begin{document}

\maketitle
\thispagestyle{empty}
\pagestyle{empty}

%%%%%%%%%%%%%%%%%%%%%%%%%%%%%%%%%%%%%%%%%%%%%%%%%%%%%%%%%%%%%%%%%%%%%%%%%%%%%%%%

\begin{abstract}
Enabling humanoid robots to exploit physical contact, rather than simply avoid collisions, is crucial for autonomy in unstructured environments. Traditional optimization-based planners struggle with contact complexity, while on-policy reinforcement learning (RL) is sample-inefficient and has limited multi-task ability. We propose a framework combining a learned world model with sampling-based Model Predictive Control (MPC), trained on a demonstration-free offline dataset to predict future outcomes in a compressed latent space. To address sparse contact rewards and sensor noise, the MPC uses a learned surrogate value function for dense, robust planning. Our single, scalable model supports contact-aware tasks, including wall support after perturbation, blocking incoming objects, and traversing height-limited arches, with improved sample efficiency and multi-task capability over on-policy RL. Deployed on a physical humanoid, our system achieves robust, real-time contact planning from proprioception and ego-centric depth images. Code and dataset are available at our website: 
    \href{https://ego-vcp.github.io/}{\textcolor{magenta}{https://ego-vcp.github.io/.}}

\end{abstract}
\section{Introduction}

Humanoids are expected to advance from dynamic locomotion~\cite{raibert1986legged,rudin2022learning} to intelligent interaction in complex, unstructured environments. Achieving this requires purposeful contact exploitation rather than simple avoidance. Effective humanoids must use their bodies to interact with the world as humans do, such as bracing against a wall for balance, blocking objects for safety, or ducking under obstacles. These contact-aware skills are essential for greater autonomy, robustness, and intelligence in robots.

Reasoning about contact remains challenging for humanoids~\cite{jenelten2024dtc, roth2025learned, liu2025discrete}. Traditional optimization-based methods~\cite{sleiman2021unified,winkler2018gait} struggle with the complexity of real-time contact scheduling and are sensitive to model inaccuracies, reducing adaptability to unforeseen situations.

Parallelized simulation~\cite{NVIDIA_Isaac_Sim} has enabled on-policy RL to succeed in robot control for quadrupeds~\cite{cheng2024extreme}, bipeds~\cite{li2025reinforcement}, and humanoids~\cite{radosavovic2024real}. However, these methods are sample-inefficient, especially with visual inputs~\cite{cheng2024extreme}, and struggle with multi-task learning.

\begin{figure}[t]
    \centering
    % \vspace{-5pt}
    \includegraphics[width=0.99\linewidth]{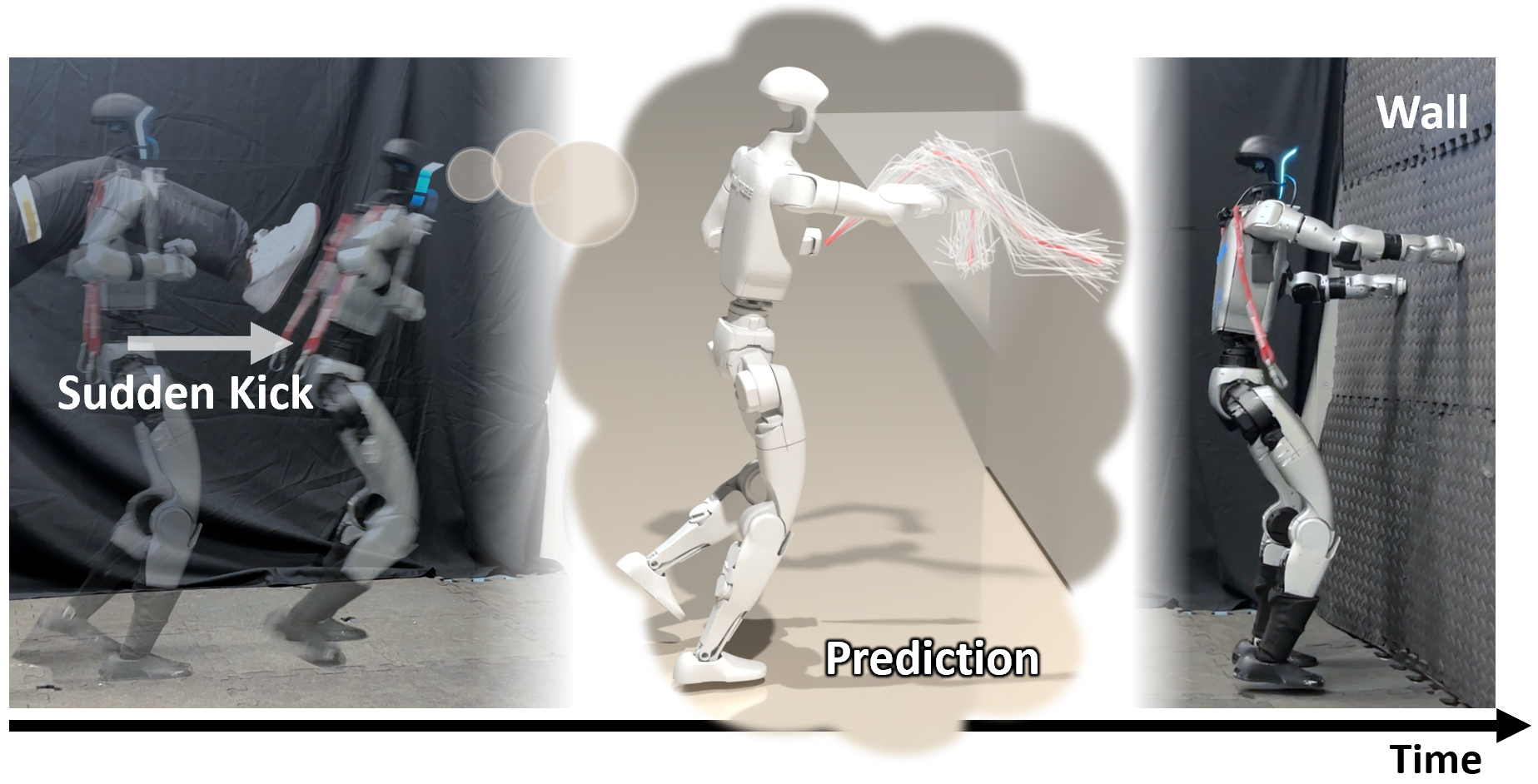}
    \vspace{-10pt}
    \caption{An illustration of our framework in the ``Support the Wall'' task. When subjected to a sudden perturbation (left), the robot uses its learned world model to predict and plan a stabilizing action within its planning horizon (center). This allows it to successfully execute the plan and brace its hands against the wall to make contact and maintain balance (right).}
    \label{fig:placeholder}
    \vspace{-10pt}
\end{figure}

We address this by integrating a learned world model with sampling-based Model Predictive Control (MPC). Our approach trains a scalable world model from a random, demonstration-free offline dataset, predicting future outcomes in the compressed latent space rather than raw pixels, and understanding action consequences. We introduce a surrogate value function to guide planning, allowing MPC to evaluate candidate action sequences efficiently. This synergy enables agile, vision-based contact planning for humanoids across tasks with improved sample efficiency and performance.

The main contributions of this work are as follows:

\begin{enumerate}[leftmargin=*]
    \item \textbf{A Scalable Visual World Model for Dynamic Robots}: We learn a visual world model that scalably captures the dynamics of diverse contact tasks, trained entirely on a demonstration-free offline dataset.
    \item \textbf{Planning from Pixels with Value-Guidance}: We introduce a sampling-based MPC framework using a learned surrogate value function to guide the planning process.
    \item \textbf{Agile and Robust Real-world Visual Contact Planning}: We validate the proposed framework on a physical humanoid robot, demonstrating multiple novel agile and robust contact planning capabilities solely from ego-centric depth images and proprioceptive feedback.
\end{enumerate}

\begin{figure*}[t]
    \centering
    \includegraphics[width=1.0\linewidth]{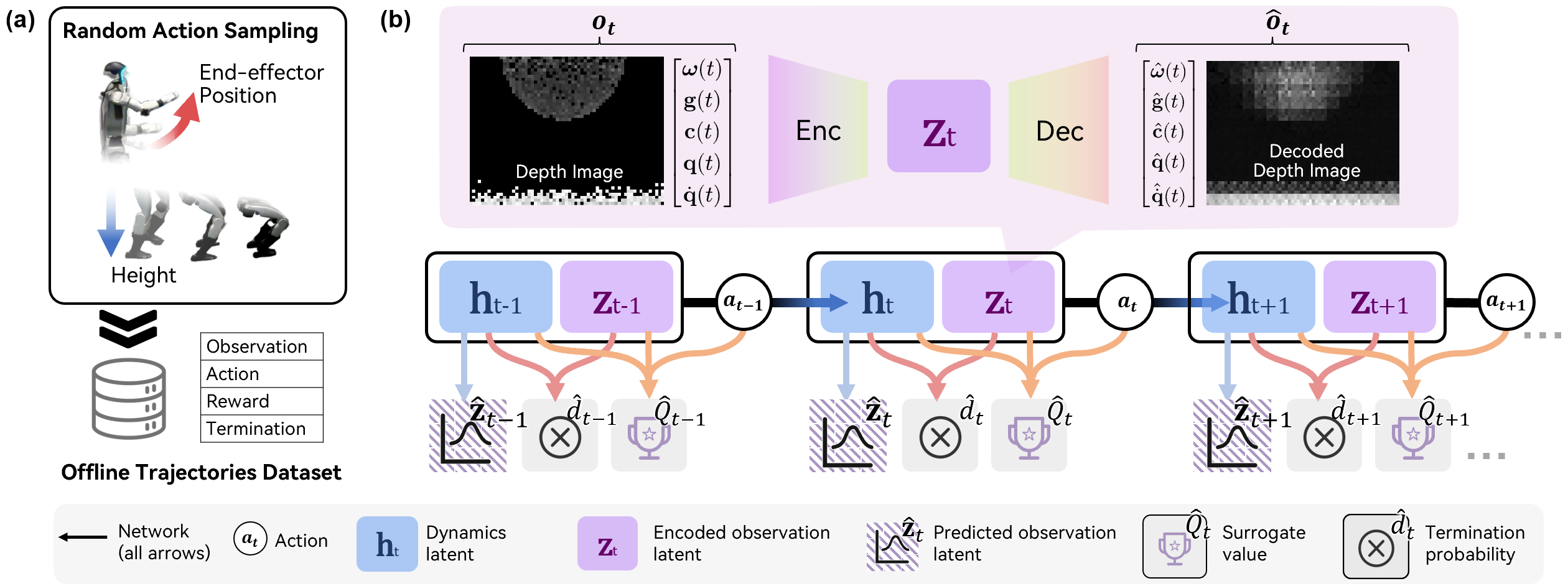}
    \caption{\textbf{World Model Training Pipeline.} The pipeline begins with the offline data collection process shown in \textbf{(a)}, where a dataset $\mathcal{D}$ of trajectories is generated by applying randomly sampled high-level actions (end-effector position $p_{ee}^{\top}$ and body height $h_{body}$) to a simulated humanoid equipped with a trained low-level policy. This dataset is then used to train the world model, as depicted in \textbf{(b)}. At each timestep $t$, an observation $o_t$, consisting of a depth image and proprioception, is encoded into an observation latent $z_t$, which is then decoded to produce a reconstruction $\hat{o}_t$. Concurrently, a recurrent network updates its latent $h_t$ based on the previous state and action. The model predicts (i) $\hat{z}_t$, a prior sample of the observation latent; (ii) $\hat{d}_t$, the termination probability; and (iii) $\hat{Q}_t$, a surrogate action-value guiding the planner in evaluating different actions. All of these predictions are optimized against the ground-truth data from the offline trajectories, enabling the model to learn both the environment's dynamics and a robust value function for planning.}
    \vspace{-0.5cm}
    \label{fig:wm}
\end{figure*}

\section{Related Work}
\subsection{Model-Based Contact Planning}

For both locomotion and manipulation, robotics is replete with contact-rich problems, made challenging by the non-smooth dynamics induced by the impact ~\cite{raibert1986legged}. Optimization-based approaches address this by explicitly modeling these physical interactions, like linearizing the complex friction model into a Linear Complementarity Problem (LCP)~\cite{stewart1996implicit}, or relaxing it into a Cone Complementarity Problem (CCP)~\cite{anitescu2006optimization}. These formulations can then be embedded within a trajectory optimization framework~\cite{sleiman2021unified,winkler2018gait}. Another prominent paradigm is Hybrid Zero Dynamics (HZD)~\cite{westervelt2003hybrid, sreenath2011compliant}, which addresses the non-smooth contact dynamics of legged locomotion by enforcing virtual constraints whose associated zero dynamics surface remains invariant through impacts. However, such model-based approaches are often hindered by model inaccuracies and high computational costs~\cite{xue2024full}, which complicate real-time deployment. Furthermore, their reliance on predefined structures, such as periodic gaits~\cite{gong2019feedback} or reference foot-end trajectories~\cite{westervelt2003hybrid}, makes it difficult to scale them to more general, aperiodic whole-body contact scenarios.

\subsection{Learning-Based Contact Planning}

Learning-based contact planning has demonstrated potential for dynamic skills~\cite{li2025reinforcement, cheng2024extreme, jenelten2024dtc, liu2025discrete, margolis2024rapid, margolis2023walk}, yet three key challenges persist. 
First, interaction capabilities are hindered by simplified 2.5D elevation maps~\cite{roth2025learned}, which fail to represent dynamic or overhanging obstacles like moving objects or archways. 
Second, sample efficiency remains a bottleneck; heavy reliance on synthetic data~\cite{NVIDIA_Isaac_Sim} makes vision-based on-policy RL computationally challenging~\cite{cheng2024extreme}, while the sparse, discontinuous nature of contact complicates exploration for model-free methods~\cite{liu2025discrete, zhang2024wococo}. 
Finally, limited multi-task adaptability prevents policies from generalizing across diverse object interactions or varying task definitions~\cite{schulman2017proximal}.

\subsection{Planning with Robotic World Models}

A world model~\cite{ha2018recurrent} provides a learned internal model of an environment, enabling robots to predict future outcomes within an abstract latent space. Integrating these models into model-based planning offers a promising trajectory for enhancing generalization and sample efficiency in contact planning~\cite{hansen2022temporal, hafner2019learning}.
While early frameworks focused on accelerating policy learning~\cite{hansen2022temporal, hafner2019learning}, contemporary world models have evolved into generative systems~\cite{assran2025v, bruce2024genie} capable of simulating complex physical causalities and dynamics. 
Parallel efforts in robotics leverage neural dynamics models to represent intricate systems~\cite{o2022neural, zhang2024adaptigraph, li2025offline, li2025robotic}, which can be integrated into sampling-based MPC~\cite{williams2017information, pan2024model, xue2024full} for adaptive control~\cite{margolislearning, roth2025learned, xiao2024anycar}.

Nonetheless, enabling robotic world models to fully generalize remains an open problem. This is especially the case for contact planning, as the underlying whole-body contact state is not directly observable and is difficult to infer and predict from partial, noisy sensory data.

\section{Methods}

Section~\ref{sec:lowlevel} first describes the \textbf{low-level controller} that executes motor control and defines the high-level planning interface used throughout the paper. We then present our \textbf{data collection} procedure for training the planner in Section~\ref{sec:data}. Finally, we detail the two core components of our high-level planner: the \textbf{architecture and training of the world model} (Section~\ref{sec:wm}), and the \textbf{value-guided sampling MPC} framework that utilizes the world model for test-time planning (Section~\ref{sec:mpc}).

\begin{figure*}
    \centering
    \includegraphics[width=0.98\linewidth]{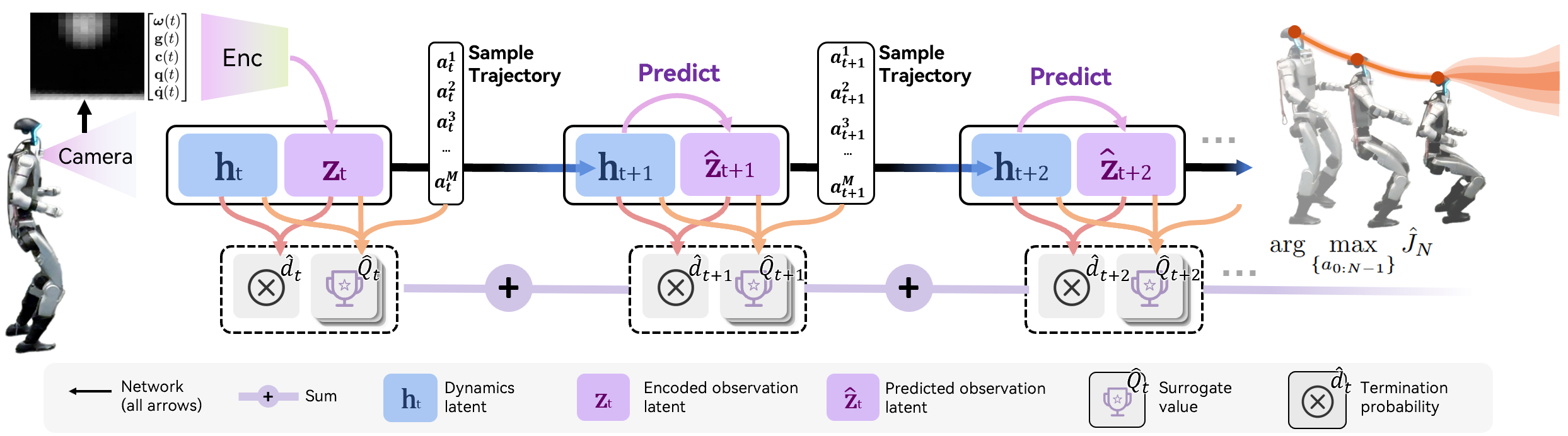}
    \caption{\textbf{Value-Guided Sampling MPC.} This figure illustrates how the trained world model is used for planning via value-guided sampling MPC. This process performs open-loop prediction to find the best action sequence starting from a single real observation. At inference time, this process begins by encoding the current observation $o_t$ into its latent state $z_t$, after which the planner samples a batch of $M=1024$ candidate action sequences over a planning horizon of $N = 4$ steps. The world model predicts the future latent state ($h_{t+k}, \hat{z}_{t+k}$) by recursively applying its learned dynamics model. At each prediction step, the surrogate value ($\hat{Q}_{t+k}$) evaluates the sampled actions, while the termination signal, $\hat{d}_{t+k}$, predicts the probability of robot failure, such as falling; if this probability exceeds a threshold of 0.9, all subsequent values for that trajectory are set to zero. The planner evaluates $M$ candidate trajectories, where the score for each trajectory is calculated by the objective function $\hat{J}_N$ in Eq.~\eqref{eq:final-obj}. This set of scored trajectories is then optimized using the Cross-Entropy Method (CEM) to find the optimal action sequence.}
    \label{fig:mpc}
    \vspace{-0.5cm}
\end{figure*}

\subsection{Low-Level Controller}
\label{sec:lowlevel}
Our framework utilizes a low-level controller capable of tracking diverse locomotion and manipulation commands. 
The controller tracks the command vector $c = [v^{\top}, p_{ee}^{\top}, h_{body}]^{\top}$, where $v$, $p_{ee}$, and $h_{body}$ denote the desired linear velocity, end-effector position, and body height, respectively. 
Its observation space is purely proprioceptive, comprising angular velocity ($\omega$), the projected gravity vector ($g$), the command ($c$), joint positions ($q$), and joint velocities ($\dot{q}$). 
The controller is trained in simulation via PPO, following established approaches~\cite{NVIDIA_Isaac_Sim, LeggedLab}.

\subsection{Data Collection}
\label{sec:data}
Once a reliable low-level controller is obtained, we collect an offline object-interaction dataset for the subsequent vision-based world model training. Our offline dataset $\mathcal{D}$ is generated by collecting trajectories $\tau$ in simulation across three object types: a ball, a wall, and an arch. At each timestep, the robot receives an observation $o_t$ including a downsampled $64 \times 48$ ego-centric depth image and proprioceptive signals, executes a randomly sampled action $a_t = \left[p_{ee}^{\top}, h_{body}\right]^{\top}$. We exclude the desired linear velocity $v$ from the planner's action space to force the robot to solve contact-rich problems through postural manipulation. In return, the robot receives the reward $r_t$ and the termination signal $d_t$ from the environment. These collected transition tuples $\{o_t, a_t, r_t, d_t\}$ are then stored in a final trajectory dataset structured as \texttt{[Batch, Time, Data]}.

At each step, we sample the finite differences of planner actions $a_t = a_{t-1} + \eta \cdot \delta $ from $\delta \sim \mathcal{U}(-1, 1)$, where $\eta$ is a scalar step that controls the magnitude of the update. This step is performed after normalizing the task space $\left[p_{ee}^{\top}, h_{body}\right]^{\top}$ of the low-level controller. $\eta$ is set to $0.32$. 
The purpose of using such a method to sample the actions is to (1) avoid using any demonstration, which is expensive to obtain for the whole-body commands of a humanoid, and (2) avoid ineffective and jittery behavior data from random sampling. For a detailed description of the process, see the illustration video on our \href{https://ego-vcp.github.io/}{\textcolor{magenta}{website}} and \href{https://github.com/HybridRobotics/Ego-VCP}{\textcolor{magenta}{code}}.

\begin{figure*}[t]
    \centering
    \includegraphics[width=1.0\linewidth]{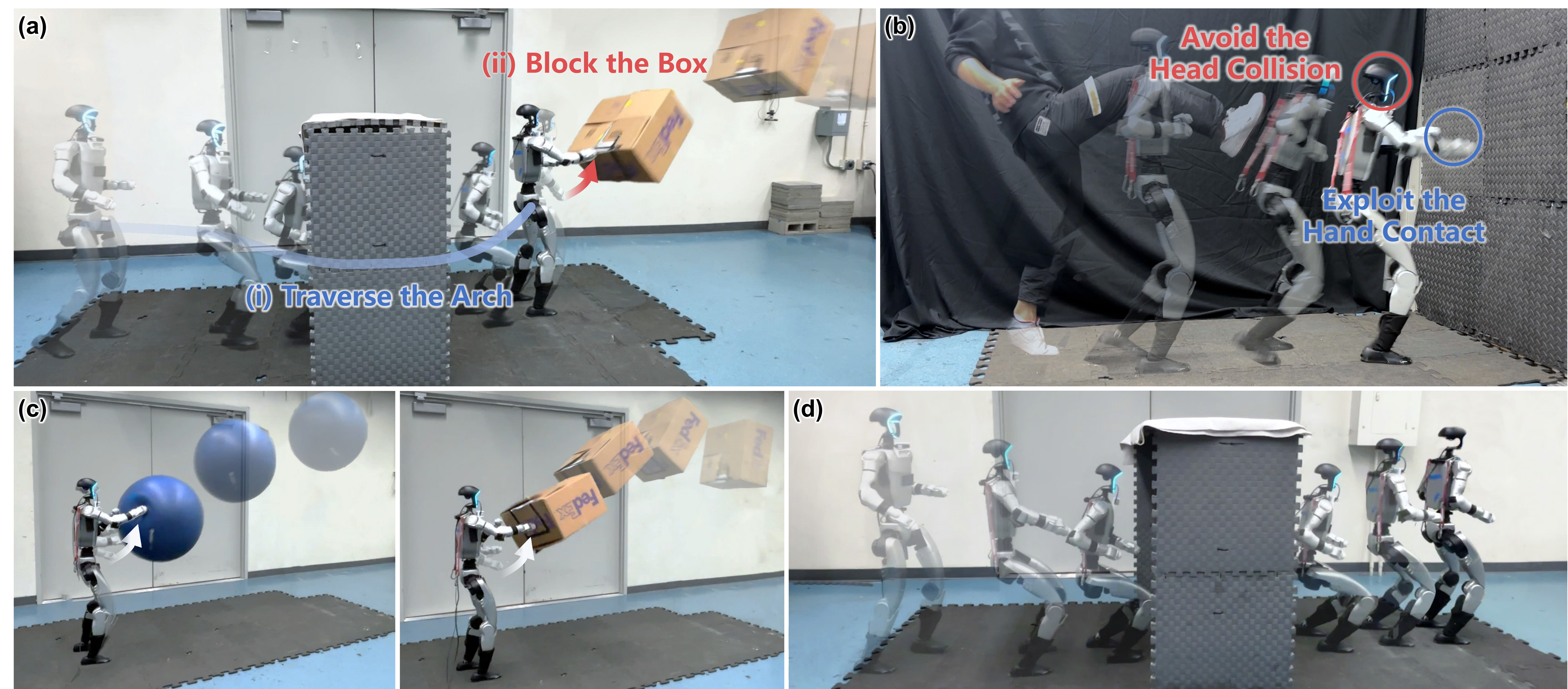}
    \caption{Real-World experiments validating the proposed framework. (a) A demonstration of sequential task execution and generalization, where the robot traverses an arch (i) and then blocks a previously unseen box (ii). (b) Support the wall to maintain balance by bracing the wall with the hands when pushed towards the wall. (c) Blocking both an in-distribution ball (with a size consistent with the training data) and an unseen box; (d) Squat and traverse an arch.}
    \label{fig:exp_all}
    \vspace{-15pt}
\end{figure*}

\subsection{Ego-Vision Humanoid World Model}
\label{sec:wm}

Prior auto-regressive models learn system dynamics by mimicking existing controllers for continuous tasks such as velocity tracking~\cite{li2025offline}. However, when applied to high-dimensional image observations, this pixel-prediction approach suffers from compounding errors over long horizons. This issue is exacerbated in \textbf{contact-aware} scenarios where defining a goal trajectory in the pixel space is often intractable.

To address this, we draw inspiration from general world models such as Dreamer~\cite{hafner2023mastering} and JEPA~\cite{assran2025v}. We focus on predicting the abstract latent states of future observations, enabling the model to capture more fundamental structures within the data. As illustrated in Fig. \ref{fig:wm}, our world model is composed of several key components detailed below.

First, the world model leverages a recurrent neural network (RNN) to maintain a deterministic dynamics latent state $h_t$. 
At each step, a stochastic latent state $z_t$, which extracts the abstract latent of the current observation, is inferred from the current observation $o_t$ and the latent state $h_{t}$. 
Similar to an autoencoder, the model is trained to reconstruct the observation $o_t$ as $\hat{o}_t$ after passing it through a bottleneck, which compels the latent state $z_t$ to encode the most salient and abstract features. For notational simplicity, we let $\phi$ denote the parameters of all world model components, with $q_{\phi}$ and $p_{\phi}$ representing the encoder and decoder, respectively. The overall process can then be expressed as:
\begin{align}
    h_t &:= f_{\phi}(h_{t-1}, z_{t-1}, a_{t-1}) \\
    z_t &\sim q_{\phi}(z_t \mid h_t, o_t) \\
    \hat{o}_t &\sim p_{\phi}(\hat{o}_t \mid h_t, z_t).
\end{align}
We model both $z_t$ and $\hat{o}_t$ as Gaussian distributions.

Furthermore, we introduce a model that estimates the stochastic latent without using the current observation: given $h_t$, it predicts a latent $\hat{z}_t \sim p_{\phi}(\hat{z}_t \mid h_t)$ that closely approximates $z_t$, thereby enabling rollouts in latent space.

Different from Dreamer~\cite{hafner2023mastering}, we need to consider an architecture that better addresses the challenges unique to robotics in the real world, such as (1) significant partial observability, (2) high sensor noise, and (3) sparse contact. These factors make it difficult to predict contact-aware rewards from the observation. Therefore, we design specialized heads that, conditioned on the latent $(h_t, z_t)$ and a candidate action $a_t$, directly estimate the expected long-term outcome. Specifically, we predict a termination probability $\hat{d}_t$ and a surrogate value $\hat{Q}_t$, which represents the expected cumulative return. This allows the robot to evaluate the potential response to different actions directly from its learned latent. 
\begin{align}
    \hat{d}_t &\sim D_{\phi}(\hat{d}_t \mid h_t, z_t), \\
    \hat{Q}_t &:= Q_\phi(h_t, z_t, a_t).
\end{align}

Our surrogate value could condition on the latent state $z_t$, allowing the robot to infer the current task context from its observations and dynamically adapt its objective. This enables us to train the model directly on a mixed dataset containing data from all tasks. We opt for a computationally efficient design consisting of a CNN for image feature extraction and MLPs for all other components.

The model is optimized by minimizing the loss as shown below. This total loss is a simple sum of three main components: a reconstruction loss ($\mathcal{L}_{\text{rec}}$), a joint-embedding predictive loss ($\mathcal{L}_{\text{jep}}$), and a Q-loss ($\mathcal{L}_{\hat{Q}}$):

\begin{equation}
\label{eq:total_loss}
\mathcal{L}_{\text{total}} = \mathcal{L}_{\text{rec}} + \mathcal{L}_{\text{jep}} + \mathcal{L}_{\hat{Q}}.
\end{equation}

The reconstruction loss, $\mathcal{L}_{\text{rec}}$, ensures the world model can extract a tight latent space of the environment. It is defined as a combination of a Negative Log-Likelihood (NLL) for the observation reconstruction ($\mathcal{L}_{\text{obs}}$) and a Binary Cross-Entropy (BCE) loss for the termination signal ($\mathcal{L}_{\text{term}}$):
\begin{equation}
\label{eq:reconstruction_loss}
\mathcal{L}_{\text{rec}} = \mathbb{E}_{{{z}}_t \sim q_{\phi}({z}_t|{h}_t, {o}_t )} \left[ \mathcal{L}_{\text{obs}} + \mathcal{L}_{\text{term}} \right].
\end{equation}
The joint-embedding predictive loss, $\mathcal{L}_{\text{jep}}$, consists of two KL divergence terms that enforce a consistent and non-collapsing latent space~\cite{chen2021exploring,hafner2023mastering}. 
\begin{align}
\label{eq:jep_loss}
\mathcal{L}_{\text{jep}} = D_{\text{KL}}\Big( \text{sg}\big(q_{\phi}({z}_t|{h}_t, {o}_t)\big) \ \big|\big| \ p_{\phi}({\hat{z}}_t|{h}_t) \Big) \nonumber \\
+ D_{\text{KL}}\Big( q_{\phi}({z}_t|{h}_t, {o}_t) \ \big|\big| \ \text{sg}\big(p_{\phi}({\hat{z}}_t|{h}_t)\big) \Big),
\end{align}
where sg is the stop-gradient operator.

The surrogate value loss, $\mathcal{L}_{\hat{Q}}$, is a mean-squared error term that trains the value function to estimate the target $Q_{\text{target}}$. We simply apply a Monte Carlo (MC) estimator here to get $Q_{\text{target}}$, and we empirically found that using an MC estimator yields more stable results in our scenario than TD-error:
\begin{equation}
\label{eq:q_loss}
\mathcal{L}_{\hat{Q}} = \mathbb{E}_{{\tau} \sim \mathcal{D}} \sum_{t}\mathbb{E}_{{z}_t \sim q_{\phi}({z}_t|{h}_t, {o}_t)}\left[ \left({Q}_{\phi}({h}_t, {z}_t, {a}_t) - Q_{\text{target}}\right)^2 \right].
\end{equation}

\begin{figure*}[t]
    \centering
    \includegraphics[width=1.0\linewidth]{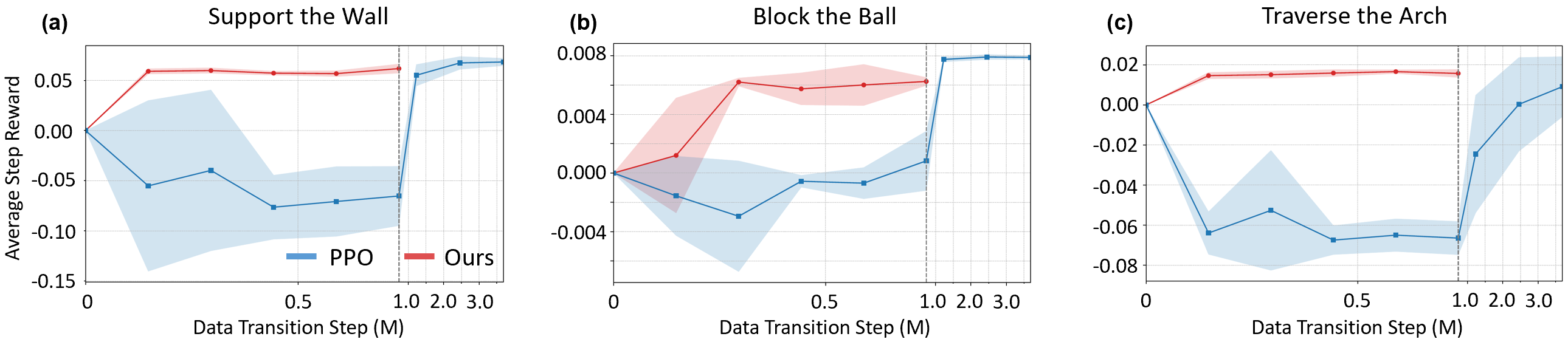}
    \caption{Sample efficiency comparison: Our method uses an offline dataset collected from random actions, while PPO collects data from environments at every iteration. The x-axis represents the number of step transitions used, while the y-axis shows the reward for each task. A greater value on the x-axis indicates a larger amount of data used, and a higher value on the y-axis signifies better performance. While our method utilizes a dataset of at most 1M steps, we continued to train PPO for a greater number of steps to determine when it could achieve comparable performance. }
    \label{fig:exp-data-effi}
    \vspace{-15pt}
\end{figure*}

\vspace{-10pt}

\subsection{Value-Guided Sampling MPC}
\label{sec:mpc}
In practical applications involving complex robotics and perception, learning a perfect observation-action value function and greedily maximizing it remains a significant challenge. This difficulty stems from two primary sources:
\begin{enumerate}[leftmargin=*]
    \item \textbf{Challenges in Offline Learning:} Finite offline datasets provide incomplete coverage, leading to unreliable value estimates for out-of-distribution actions.
    \item \textbf{Partial Observability and Physical Non-idealities:} Robotic systems suffer from partial observability, as the full state, such as contact forces, is not directly measured and is subject to sensor noise and action delays, both of which degrade value estimation.
\end{enumerate}
To address this, we introduce a Value-Guided Sampling MPC framework. This approach explicitly treats the learned value function not as an optimal oracle, but as a powerful, albeit imperfect, heuristic to guide a robust, receding-horizon planning process.

Consider a standard MDP with state $s_t$ and action $a_t$ for the analysis of the variance reduction of our proposed method. When the perfect $Q$ is available, we can leverage the Bellman principle to obtain the optimal control policy:
\begin{equation}
\label{eq:greedy_policy}
\pi(s_t) = \arg\max_{a_t} {Q}(s_t, a_t). 
\end{equation}
However, in practice, only the estimation $\hat{Q}$ is available, which may have a large variance. The imperfection of $\hat{Q}$ may lead to performance degradation. To mitigate this issue, we consider an $N$-step surrogate optimization:
\begin{equation}
\begin{aligned}
    \hat{\pi}(s_t) = \arg\max_{\{a_{t:t+N-1}\}}& \hat{J}_N := \frac{1}{N}\sum_{k=0}^{N-1} \hat{Q}(s_{t+k}, a_{t+k}) \\
    \mathrm{s.t. } & \quad s_{t+k+1} = f_{\phi}(s_{t+k}, a_{t+k})
\end{aligned}
\label{eq:obj}
\end{equation}
% \junfeng{
% Should be:
% \begin{equation}
% \begin{aligned}
%     \hat{\pi}(s_t) &= \arg\max_{\{a_{0:N-1}\}} \hat{J}_N := \frac{1}{N}\sum_{t=0}^{N-1} \hat{Q}(s_t, a_t) \\
%     \mathrm{s.t }& \quad s_{t+1} = f(s_t, a_t)
% \end{aligned}
% \end{equation}
% }
By the definition of the $Q$ function, $\hat{\pi}(\cdot)$ and $\pi$ are obtained by solving two different optimal control problems, thus they are in general not identical. Despite the biases, we show that the surrogate objective can reduce the variance. 

Consider the residual $\epsilon_{t+k}$ at time $(t+k)$ as:
\begin{equation}
\epsilon_{k}:=\hat{Q}(s_{t+k}, a_{t+k}) - Q(s_{t+k}, a_{t+k}).  
\end{equation}
For shorthand notation, we use the subscript $(k)$ to denote the quantity at time step $t+k$. Consider the surrogate objective:
\begin{equation}
    \frac{1}{N}\sum_{k=0}^{N-1} \hat{Q}_k = \frac{1}{N}\sum^{N-1}_{k=0}\left(Q_k + \epsilon_{k}\right).
\end{equation}
% We assume that the variance of $\epsilon_Q$ is bounded by $\operatorname{Var}[\epsilon_Q] \le V$.
% As we use Monte Carlo estimation to compute $\hat{Q}$ as an unbiased estimation, the mean of $\epsilon_t$ is zero, and thus the variance on the RHS can be obtained by
Let $J_N:=\frac{1}{N}\sum_{k=0}^{N-1}Q_k$. As we use Monte Carlo returns to form $Q_{target}$ and train $\hat{Q}$ via regression, we have:
\begin{equation}
\begin{aligned}
    \mathbb{E}[(J_N-\hat{J}_N)^2]&=\mathbb{E}[(\frac{1}{N}\sum^{N-1}_{k=0}\epsilon_{k})^2] \\
    &= \frac{1}{N^2}\sum_{(i, j)} \operatorname{Cov}[\epsilon_{ i}, \epsilon_{j}] + \bar{\epsilon}^2. \\
\end{aligned}
\end{equation}
with $\bar{\epsilon}:=\mathbb{E}[\epsilon_k]$ denoting the bias introduced by estimating $Q_k$ corresponding to the infinite horizon MDP using truncated trajectories. 
We assume that the variance of $\epsilon_k, \forall k,$ satisfies: \begin{equation}
    0 \le \sigma_{\min} \le \sqrt{\operatorname{Var}[\epsilon_{ k}]} \le \sigma_{\max}, \tag{Bounded Variance}
\end{equation}
and the correlation is bounded by $\rho < 1$, such that $\forall i, j$ \begin{equation}
    |\operatorname{Cov}[\epsilon_i, \epsilon_j]| \le \rho \sqrt{\operatorname{Var}[\epsilon_i] \operatorname{Var}[\epsilon_j]}.   \tag{Bounded Correlation}
\end{equation}

The upper bound on the variance term is achieved if all steps are positively correlated: 
\begin{equation}
\begin{aligned}
        % \operatorname{Var}[\frac{1}{N}\sum^{N-1}_{t=0}\epsilon_{ t}]  & 
        \mathbb{E}[(J_N-\hat{J}_N)^2] - \bar{\epsilon}^2 &\le \frac{N+\rho(N^2-N)}{N^2} \sigma^2_{\max}  \\
        & =: V_{\operatorname{ub}}.
        % \frac{1}{N^2}N^2 \max_{(i, j)}\operatorname{Cov}[\epsilon_{ i}, \epsilon_{ j}] = V_{\max}.
\end{aligned}
\end{equation}
For the lower bound, similarly, we have:
\begin{equation}
\begin{aligned}
                % \operatorname{Var}[\frac{1}{N}\sum^{N-1}_{t=0}\epsilon_{ t}]  & 
                \mathbb{E}[(J_N-\hat{J}_N)^2] - \bar{\epsilon}^2 & \ge \max \{ \frac{N \sigma^2_{\min} - \rho(N^2-N)\sigma^2_{\max}}{N^2}, 0 \} \\ & =: V_{\operatorname{lb}}.
\end{aligned}
\end{equation}
Thus, we have the limit when $\rho \rightarrow 0: V_{\operatorname{ub}} \rightarrow \frac{\sigma^2_{\max}}{N}, V_{\operatorname{lb}} \rightarrow \frac{\sigma^2_{\min}}{N}$. As we compute $\hat{Q}$ from an offline dataset generated by random actions in each step, the correlation between $\hat{Q}_i$ from different time steps is weaker than by sampling using a state feedback policy. Thus, it is possible that $\rho$ is small and $\hat{J}_N$ has a substantially lower variance than $\hat{Q}$. Although the surrogate objective does not preserve a local optimum, if the variance of $\hat{Q}$ dominates, this strategy can significantly improve the performance.

Based on the above analysis, we apply the surrogate \textbf{objective function} $\hat{J}_N$ from ~\eqref{eq:obj} using the latent $h_t$ and $\hat{z}_t$ as the representation of the robot states. The optimal sequence, denoted $A_t^* = \{a_t^*, a_{t+1}^*, ..., a_{t+N-1}^*\}$, is the one that maximizes our surrogate objective $\hat{J}_N$:
\begin{equation}
\begin{aligned}
    A_t^* = \arg\max_{A_t} &\ \ \frac{1}{N}\sum_{k=0}^{N-1} Q_\phi(h_{t+k}, \hat{z}_{t+k}, a_{t+k}), \\
    \mathrm{s.t.} & \quad h_{t+k+1} = f_{\phi}(h_{t+k}, \hat{z}_{t+k}, a_{t+k}), \\
                 & \quad \hat{z}_{t+k} \sim p_{\phi}(\hat{z}_{t+k} \mid h_{t+k}).
\label{eq:final-obj}
\end{aligned}
\end{equation}
As shown in Fig. \ref{fig:mpc}, the framework’s memory is maintained in two latent states: a dynamics latent state $h_t$ and a current observation latent state $z_t$. The $h_t$ is computed by the RNN from the previous latent ($h_{t-1}, z_{t-1}$) and the last action ($a_{t-1}$). The $z_t$ is then generated in one of two ways: when an observation $o_t$ is available, $z_t$ is inferred using both the latent state $h_t$ and the observation $o_t$; for future predictions, it is generated from the latent state $h_t$ alone. Our world model also predicts the probability of robot failure, $\hat{d}_{t}$, such as falling; if this probability exceeds a threshold of 0.9, all subsequent value estimates for that trajectory are set to zero.

We use the Cross-Entropy Method (CEM) to find the optimal action sequence $A_t^*$. Once identified, only the first action $a_t^*$ is executed. This iterative re-planning allows the robot to continuously incorporate feedback from the environment, enabling it to react to disturbances and correct for model inaccuracies in real-time. From Table~\ref{tab:results}, we select a planning horizon $N=4$ as our default, as it provides the best overall performance across all tasks.

\begin{figure*}
    \centering
    \includegraphics[width=0.9\linewidth]{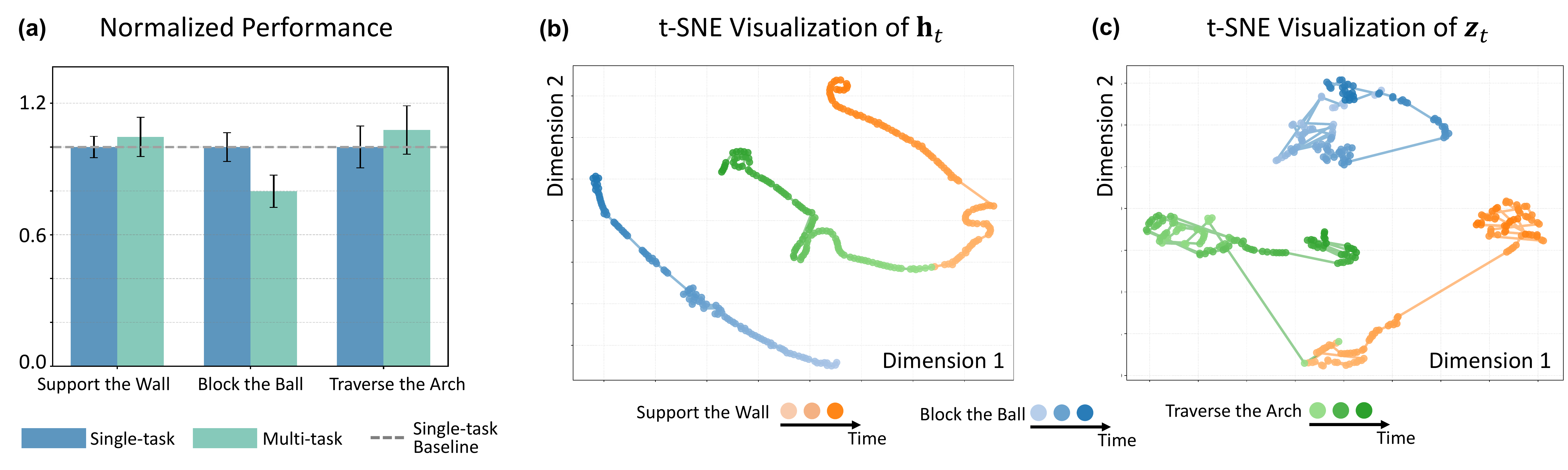}
    \caption{Multi-task performance and latent space visualization. (a) A single model trained jointly on all tasks achieves comparable normalized performance to specialized single-task models. (b-c) The t-SNE visualizations reveal a clear separation of tasks in the latent spaces. As envisioned by our design, the latent $h_t$ primarily represents dynamics, showing significant evolution over time, while the latent $z_t$ provides a more compressed representation of the current environmental observation. }
    \label{fig:multi}
    \vspace{-15pt}
\end{figure*}

\section{Experiments}

Our experimental platform is the Unitree G1 humanoid robot, equipped with a RealSense D435i camera. All quantitative analyses, including ablation studies and baseline comparisons, are conducted in a controlled simulation environment. All comparisons are conducted using a consistent set of training epochs and hyperparameters. The mean and standard deviation are then computed from ten independent trials across three different random seeds. We subsequently validate our approach with real-time experiments on a physical robot. We designed three core tasks to evaluate the model's ability to plan and execute diverse contact-rich behaviors, including \textbf{exploiting contact for stability} and \textbf{avoiding contact for safety}.

\paragraph{Task}(1)~\textbf{Support the Wall}: the robot must resist external disturbances by stabilizing itself only through supportive hand contact;
(2)~\textbf{Block the Ball}: the robot must intercept a flying object only with defensive hand contact;
(3)~\textbf{Traverse the Arch}: the robot must pass through a low-clearance arch while avoiding unintended head contact.

\paragraph{Baselines} 
We compare our method with these baseline methods:
(1) \textbf{PPO}: an implementation in \cite{rudin2022learning}.
(2) \textbf{ARWM}: replaces our framework with auto-regressive prediction training like~\cite{li2025offline, li2025robotic}. 
(3) \textbf{Rew-MPC}: replaces objective function Eq.~\eqref{eq:final-obj} with $\sum_{k=0}^{N-1} \gamma^k\hat{r}_k$ from PlaNet~\cite{hafner2019learning}.
(4) \textbf{TD-MPC}: replaces objective function Eq.~\eqref{eq:final-obj} with  $\sum_{k=0}^{N-1} \gamma^k\hat{r}_k + \gamma^N\hat{Q}$ from TD-MPC~\cite{hansen2022temporal}.

\begin{table}
\centering
\caption{Single-Task Reward Evaluation of Our Method and Baselines: we analyze the influence of three key design choices on performance: the planning horizon $N$, the world model training methodology, and the objective function.}
\label{tab:results}
\setlength{\tabcolsep}{3pt}
\begin{tabular}{llccc}
\toprule
\multicolumn{2}{l}{\textbf{METHOD}} & \textbf{Reward:Wall} $\uparrow$ & \textbf{Reward:Ball} $\uparrow$ & \textbf{Reward:Arch} $\uparrow$ \\
\midrule
\multicolumn{5}{c}{\textbf{HORIZON $N$}} \\
\midrule
 & Ours, N=1 & $0.0557 \pm 0.0047$ & $-0.0066 \pm 0.0050$ & $-0.0396 \pm 0.0121$ \\
 & Ours, N=2 & $0.0607 \pm 0.0023$ & $0.0056 \pm 0.0003$ & $0.0154 \pm 0.0011$ \\
 & Ours, N=3 & $0.0611 \pm 0.0025$ & $0.0059 \pm 0.0003$ & $0.0156 \pm 0.0011$ \\
 & Ours, N=4 & $0.0614 \pm 0.0027$ & $\mathbf{0.0061} \pm 0.0003$ & $\mathbf{0.0157} \pm 0.0015$ \\
 & Ours, N=5 & $0.0598 \pm 0.0049$ & $0.0058 \pm 0.0012$ & $0.0144 \pm 0.0062$ \\
 & Ours, N=6 & $\mathbf{0.0617} \pm 0.0031$ & $0.0053 \pm 0.0020$ & $0.0115 \pm 0.0099$ \\
\midrule
\multicolumn{5}{c}{\textbf{WORLD MODEL}} \\
\midrule
 & ARWM & $0.0609 \pm 0.0047$ & $0.0039 \pm 0.0033$ & $-0.0018 \pm 0.0183$ \\
\midrule
\multicolumn{5}{c}{\textbf{OBJECTIVE FUNCTION}} \\
\midrule
 & Rew-MPC & $0.0302 \pm 0.0204$ & $-0.0033 \pm 0.0044$ & $-0.0211 \pm 0.0092$ \\
 & TD-MPC & $\mathbf{0.0699} \pm 0.0035$ & $-0.0016 \pm 0.0047$ & $0.0145 \pm 0.0005$ \\
\bottomrule
\end{tabular}
\vspace{-5mm}
\end{table}

\subsection{Advantages of the Use of Offline Data}

\textbf{Sample Efficiency In Single-Task: } We first compare our method against PPO implemented in \cite{rudin2022learning}, an online on-policy RL algorithm that remains the dominant training method in the legged robotics domain. A key distinction is that PPO requires continuous interaction with the environment, whereas our approach is fully offline, trained from a fixed, demonstration-free dataset without any environment interaction. We do not compare against off-policy methods such as SAC and its variants, which, although more sample-efficient than on-policy approaches, are less commonly applied to humanoid robots in real-world settings due to their limited scalability to complex, high-DoF dynamic control.

As shown in Fig. \ref{fig:exp-data-effi}, in three contact-reward-dominant tasks, our method completes the tasks using only 0.5M data steps. In contrast, PPO requires a significantly larger amount of data, especially in simulations that necessitate visual rendering, where it consumes considerably more time. While PPO can quickly match our efficiency in tasks with simple visual features and a stationary robot, such as ``Block the Ball," our method achieves better performance on tasks with more complex visual representations and significant changes in the robot's viewpoint. For instance, in the ``Traverse the Arch" task, where the robot’s perspective changes dramatically between standing and squatting positions, our method substantially outperforms PPO.

\textbf{Multi-Task Capability: }In addition to sampling efficiency on the single task, we qualitatively analyze the challenges of applying online RL like PPO to a multi-task setting:
\begin{itemize}[leftmargin=*]
    \item \textbf{Reward Engineering:} online RL requires either (1) a unified reward function across all tasks, which is difficult to design, or (2) substantial engineering effort to implement complex logic for task switching and conditional rewards.
    \item \textbf{Catastrophic Forgetting \& Curriculum Design:} online RL is prone to catastrophic forgetting. Mitigating this requires a carefully designed task sampling curriculum, the complexity of which grows as more tasks are added.
\end{itemize}
Our approach, which leverages offline data, circumvents these challenges by learning directly from a mixed dataset, enabling effective and scalable multi-task training. The result of our method in the multi-task setting is shown in Sec. \ref{sec:exp-multi}.

\begin{figure*}[t]
    \centering
    \includegraphics[width=0.9\linewidth]{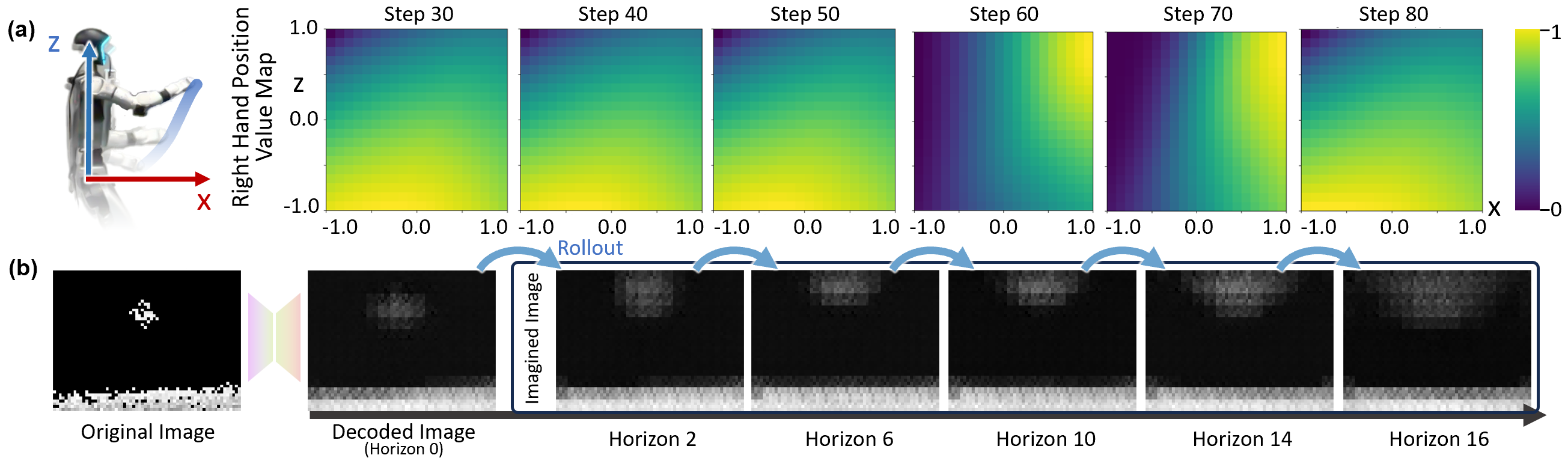}
    \caption{Visualization of the world model's prediction and planning process during the ``Block the Ball'' task. (a) The evolution of the Q-value map for the hand's position in the X-Z plane. As the task unfolds (from Step 30 to 80, with the ball appearing at Step 60), the model dynamically updates its value estimates, with the high-value region (yellow) indicating the optimal location to intercept the object. (b) An open-loop prediction of future observations. Given an initial image, the model first reconstructs it (Horizon 0) and then generates a sequence of future frames (Horizon 2-16) by decoding its predicted latent states, visualizing its anticipation of the ball's trajectory. It is worth noting that while our planner uses a shorter \textbf{4-step horizon}, this visualization is for demonstrating the long-horizon physical intuition.}
    \label{fig:value}
    \vspace{-15pt}
\end{figure*}

\subsection{Analysis of Key Design Choices}

To demonstrate the necessity of our design, as shown in the Table \ref{tab:results}, we found that greedily maximizing the value (i.e., the horizon-1 case) is infeasible. This approach causes the robot to be myopic, favoring the maintenance of its default position and ignoring future contacts. We observe that longer horizons (e.g., $N{=}6$) degrade performance, likely because bias dominates from longer-term prediction (see Sec.~\ref{sec:mpc}), whereas $N{=}4$ strikes a bias–variance sweet spot.

We also find that incorporating autoregressive prediction in our method (the ARWM baseline in Table~\ref{tab:results}) is not necessary and can even be detrimental to value estimation in offline RL, as it overemphasizes precise prediction, which leads to value function overfitting. 

As shown in the Table \ref{tab:results}, Rew-MPC, which uses reward as the objective function, yielded suboptimal results due to partial observability, noise, action lag, and sparse contact, which make rewards difficult to predict. And TD-MPC, which uses TD-target to evaluate trajectories, also produces unstable results. We attribute this to TD-error methods converging to deceptive solutions, where low TD error might mask highly inaccurate value estimates. As argued in~\cite{fujimoto2022should}, this phenomenon is caused by bias cancellation and the existence of infinitely many suboptimal solutions that satisfy the Bellman equation on an incomplete offline dataset.

\subsection{Multi-Task Planning with Unified World Model}
\label{sec:exp-multi}

To evaluate the multi-task capability, we trained a single model on a combined dataset from all tasks and compared it to the single-task models from Table~\ref{tab:results}. As shown in Fig.~\ref{fig:multi}, the multi-task model achieves improved performance on two of the three tasks, with a minor drop in the ``Block the Ball" task, which we attribute to its smaller reward scale. 

To understand how this is achieved, we visualized the latent spaces using t-SNE. The visualizations reveal that our model learns to form distinct clusters for each task. The latent state $h_t$ shows significant temporal evolution, confirming that the model learns to encode the unique latent dynamics for each task from the mixed data.

\subsection{Model Interpretation and Visualization on Prediction}

We provide visualizations in Fig.~\ref{fig:value} that offer insight into the internal decision-making process of our framework. Specifically, we analyze this process on two levels: first, whether our model has learned a genuine understanding of the environment's dynamics, and second, how it leverages this understanding for its decision-making process.

\textbf{Physical Intuition:} To assess whether the learned model captures task-relevant physical properties, rather than collapsing to trivial solutions, we apply long-horizon open-loop rollouts for verification. 
As shown in Fig.~\ref{fig:value}~(b), the rollout preserves the parabolic motion of the ball over 16 steps, indicating that the latent states retain physically meaningful structure.
Importantly, exhibiting long-horizon physical intuition in open-loop rollouts does not imply that increasing the MPC planning horizon will necessarily improve control performance. In sampling-based MPC, longer horizons typically induce a more challenging optimization landscape and exacerbate model bias, so a short 4-step receding horizon empirically provides a practical sweet spot, and we replan frequently.

\textbf{Contact-Directed Planning:} Fig.~\ref{fig:value}~(a) visualizes how the model leverages its predictions for planning by showing the evolution of the objective function value in Eq.~\eqref{eq:obj} for the hand's target position. Early in the task (e.g., Step 30), the value map consistently encourages the hand to stay near a natural, energy-efficient default position. However, as the ball approaches and the plan solidifies (e.g., Step 60-70), a high-value region (yellow) emerges and sharpens, decisively guiding the robot's hand toward the optimal contact point. This dynamic evolution of the value map showcases the model's contact-directed reasoning, effectively forming an interpretable plan to achieve its objective.

\subsection{Real-World Validation}
We deployed our method for real-time experiments on Unitree G1 with 25 Hz real-time planning and evaluated a batch of 1024 action trajectories with a planning horizon of 4 steps at each timestep. The desired base velocity $v$ was controlled by a human operator.

Our real-world deployments, shown in Fig.~\ref{fig:exp_all}, included both single-task and multi-task models, both of which proved capable of completing their assigned tasks. The policy also demonstrated the ability to generalize to out-of-distribution (OOD) scenarios, such as blocking a previously unseen box. These experiments validate that our method can achieve agile and robust vision-based control. Crucially, the learned policy exhibits reactive, context-dependent behavior rather than overfitting to a single action pattern. For instance, in the ``Support the Wall'' task, the robot only braces its hands against the wall when actively disturbed and returns to a neutral stance once balance is recovered.

% \input{sec/6_lim}

% \section{Discussion and Limitations}

% \textbf{Limitations 1:} For long-horizon tasks such as pick and place, this method still needs demonstration.

% \textbf{Limitations 2:}
% Though \emph{value-guided surrogate loss} can potentially lower the variance of the estimation value, it generally does not preserve the local optimum of the original optimal control problem. The temporal correlation between $\hat{Q}$ along a trajectory should be calibrated in the future to have a tighter quantification of the variance reduction. 

\section{Conclusion}
By integrating a scalable ego-centric visual world model with value-guided sampling-based MPC, we demonstrate that humanoid robots can efficiently and robustly learn agile, contact-rich behaviors from offline, demonstration-free data, advancing data-efficient, vision-based planning for real-world robotic interaction.
% \newpage

% \section*{APPENDIX}

\section*{ACKNOWLEDGMENT}

This work was supported in part by NSF Grant CMMI-1944722 and by the Robotics and AI Institute. M. Ghaffari receives support from AFOSR MURI FA9550-23-1-0400. Z. Li was funded in part by the InnoHK initiative of the Innovation and Technology Commission of the Hong Kong Special Administrative Region Government via the Hong Kong Centre for Logistics Robotics. The authors would like to thank Jiaze Cai and Yen-Jen Wang for their help with experiments. We are also grateful to Bike Zhang, Fangchen Liu, Chaoyi Pan, Junfeng Long, and Yiyang Shao for their valuable discussions. This work was done during H. Liu's visit to UC Berkeley.

%%%%%%%%%%%%%%%%%%%%%%%%%%%%%%%%%%%%%%%%%%%%%%%%%%%%%%%%%%%%%%%%%%%%%%%%%%%%%%%%

% \bibliographystyle{IEEEtran}
% \bibliographystyle{unsrtnat}% {plainnat}
% \bibliography{references}

% {\small
% \balance
% \bibliographystyle{unsrtnat}% {plainnat}
% % \bibliography{references}
% }
\balance
\printbibliography

\end{document}